\documentclass{article}[]
\usepackage{epsfig}
\usepackage{psfig}
\usepackage{graphicx}
\usepackage{url}
\usepackage{booktabs}
\usepackage{caption}
\usepackage{subcaption}

\begin{document}

\title{{\bf Email Classification into Relevant Category Using Neural Networks}\\ }

\author{Deepak Kumar Gupta \& Shruti Goyal\\ \vspace{0.5cm}\\
 Co-Founders: Reckon Analytics \vspace{0.5cm}\\ \emph{deepak@reckonanalytics.com \& shruti@reckonanalytics.com}
 }
\date{July 10, 2017}
\maketitle
\begin{abstract}
In the real world, many online shopping websites or service provider have single email-id where customers can send their query, concern etc. At the back-end service provider receive million of emails every week, how they can identify which email is belonged of a particular department? This paper presents an artificial neural network (ANN) model that is used to solve this problem and experiments are carried out on user personal Gmail emails datasets. This problem can be generalised as typical Text Classification or Categorization \cite{sebastiani2002machine}.

\textbf{Keywords}: Artificial Neural Network, Email Classification, Natural Computing, Text Categorization
\end{abstract}

\section{Introduction}
\label{intro}

Electronic mail or e-mail is a method of electronic communication between two or more users using the Internet. Nowadays emails are not just used for communication but also used for managing the task, solving customer queries. Email Classification or Categorization has been inspired from the text categorization in machine learning (Supervised) and now email classification has been adopted in different variations such as categorising emails into pre-defined folder, blocking spam email, identifying tone of consumer from email content etc.\\

Latest email application and the service provider such as Gmail, Outlook allow the user a simple method of filtering incoming emails based on the email subject, keywords in the body, this method best suitable for personal work or home users, which means one need to create keyword-based rules to filter emails into different folders. Implementing or Creating these rules manually in email software can be difficult if one wants to categorise each incoming emails. X. Carreras and L. Marquez \cite{carreras2001boosting} noted that most users waste a large amount of time in managing their emails or they simply do not prefer to create keyword-based rules to filter email inbox\\

Today, in the world of big data, the volume of emails growing so fast. As per Radicati \cite{levenstein2013email} , in 2016, there was 2672 million email user who exchanged about 215.3 billion emails per day. It is estimated by The Radicati Group \cite{levenstein2013email}  that email database will grow by 4.7\% \\

Consider a large eCommerce website which has about the active customer base of 200 million and supposes at least 10\% customer make a purchase every month. Gaint eCommerce like Amazon, eBay etc generally has a common email address (cs@amazon.com) for all kind of queries. It means that an eCommerce company must be getting about 100000 emails every month (considering 10\% customer do write emails), therefore a company requires big database to store all emails and a system which can automatically identify/classify an email into correct department categories such as Refunds, Shipping, Quality Issues etc. \\

A customer support manager who is responsible for assigning thousands of emails to respective teams so that quick solutions and service can be provided to the customer. Imagine how much time one has to spend if a company gets millions of e-mails during Boxing Day or Black Friday sale. Companies do need a system which can automatically classify or assign a label to an email. However, To provide customer support using email communication channel one need a model or system which learns from the previous dataset and categorise new emails with higher accuracy is very much desired by companies.\\

\begin{figure}[!htb]
\centering
\includegraphics[width=0.50\textwidth]{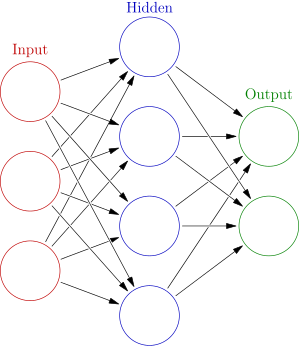}
\caption{Basic Neural Network.}
\label{fig:anns}
\end{figure}

This study investigates and carried out experiments to find out how artificial neural networks algorithm can be utilised for email classification.\\

\section{Artificial Neural Networks}

Artificial Neural Networks (ANNs) can resemble with the human brain. The key element of a neural network is a general model of a Neuron Perceptron Fig.~\ref{fig:anns}. A neural network consists of a set of neurons and each neuron is connected to one or more neurons in a direct manner.  \\

\begin{figure}[!htb]
\centering
\includegraphics[width=1\textwidth]{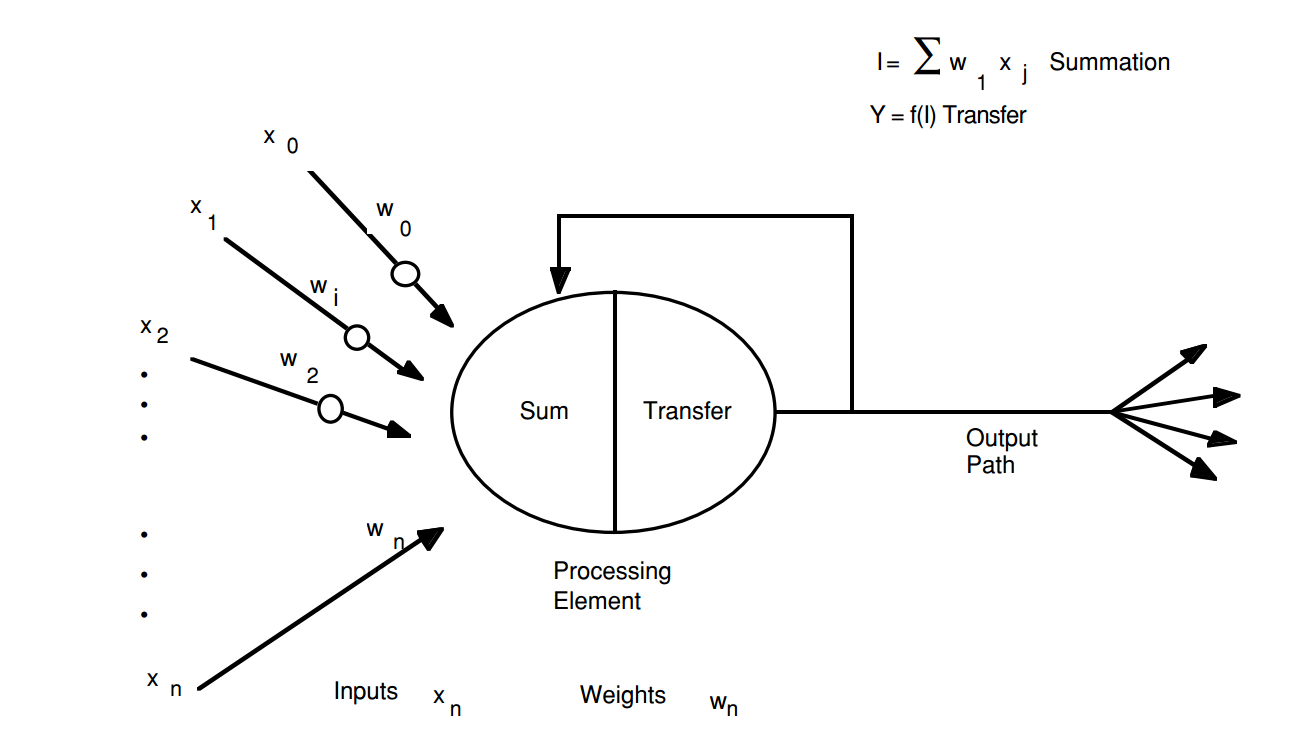}
\caption{Neural Network Architecture}
\label{fig:neural}
\end{figure}

Anderson, Dave and McNeill, George\cite{anderson1992artificial} noted that an artificial neural network consists of multiple inputs which can be represented with the help of symbol, $x(n)$ Fig.~\ref{fig:neural} and these inputs directly fed into the network. Information from inputs is weighted $w(n)$ Fig.~\ref{fig:neural}  before sending to next level layers i.e hidden layers depending on the number of hidden layers in a network. Connection weight of each input is summed and then directly fed via a transfer function to produce final output i.e. classification of data. \\

There are three types of learning for an ANNs: Supervised, Unsupervised and Reinforced. Supervised learning is more commonly used for training a neural network for a given dataset. One can train the perceptron with supervised learning in ANNs by calibrating the inputs weights. For supervised learning, training dataset already has predefined labels or category for given input weights. Each training dataset is fed into perceptron which performs some computation and then generates an output. The output result is matched against predefined class/label, no input weight adjusted if it's a match otherwise input weight slightly modified according to the expected final results. The process is repeated number of time so that model can be trained with higher accuracy. As per \cite{shao2011comparison}  "The most appropriate point to stop training may be the point at which the reduction of Mean Square Error (MSE) becomes marginal." \\

A propositional algorithm developed by Cohen \cite{cohen1996learning} called RIPPER to categorising emails into folders based on "keyword-spotting rules". Cohen also said that keyword spotting rules are easy to create, update and reuse. On the other hand, Sahami \cite{sahami1998bayesian} did classification of Spam emails using a bag of words by applying the naive Bayesian algorithm.

\section{Methodology}

\subsection{About Dataset}

To perform experiments on neural network for email classification, personal Gmail inbox data has been imported. In this data, each email has been assigned a pre-defined class/category using \textit{\textbf{Gmail\textsuperscript{TM} label}} feature and dataset has 608 emails.

\begin{table}[!htb]
\caption{Label Email Count Breakdown}

\centering
\begin{tabular}{l|c}  % centered columns
\hline                      %inserts double horizontal lines
{\bf Label}&{\bf Count}\\ \hline
bvp\_                                & 102             \\
corprova2011                         & 238             \\
deepak@gmail.com           & 73              \\
Inbox                                & 47              \\
gupta@live.com                   & 21              \\
Imagic                               & 91              \\
Placement                            & 8               \\
Friends                               & 17              \\
Jobs                                  & 11              \\ \hline
\bf Total emails                     & \bf 608         \\ \hline
\end{tabular}

\label{emailcount}
\end{table}

\subsection{Model Setup}
This experiment has been carried out on a neural network which is build using Keras \cite{chollet2015keras} as a back-end, as Keras provide a playground that facilitates easy and fast implementation using Python to carry out deep learning experiment in a jiffy. Keras convert emails into a numeric matrix by assigning a rank based on the number of times a word appeared, thereafter converting the number into vectors which represent the label of each email. A neural network can be trained by consuming this data with LSTM ( Long Short Term Memory) to correct categorise new emails. But, by extracting best and bad features for each email, neural net accuracy improved. To extract features from the text, Keras tokenizer class allow us to do so.

\begin{figure}[!htb]
\centering
\begin{verbatim}
\begin{verbatim}
Label word breakdown:
	Bank:4
	bvp_:7441
	Sent:1561
	Unread:6810
	corprova2011:10932
	deepak@gmail.com:3880
	Inbox:4244
	Starred:60
	gupta@live.com:1093
	Google:347
	Imagic:3982
	MyUnplugged:34
	Placement:401
	Friends:228
	Jobs:1202
Total word count: 42219
\end{verbatim}
\caption{Output: Word Count for each Label in dataset }%
\label{code:wordcount}%
\end{figure}

\section{Experiments \& Results}

Experimental data used is from first author personal Gmail inbox. Initially, the dataset has been cleaned and synthesised for the consumption of ANNs and labels which do not have enough number of emails has been removed. Such emails will not contribute much in training of the model. The model has been trained using 548 (90.13\%) emails and 60 (9.86\%) emails used for validating the model. Two experiments carried out to measure the performance of the model for different input parameters and the number of words selection.

\subsection{Experiment \#1}

In this experiment number of words, the ratio has been fixed while the number of hidden layers changed to test the performance of the neural network with an increase in the number of hidden layers. When $HN = 1$ (Number of Hidden Layers), the accuracy achieved was 33\%, which is very poor but when hidden layers increased linearly, it is observed that accuracy of algorithm increased to 85\% when HN = 100.  \\

\begin{figure}[!htb]
\centering
\includegraphics[width=1\textwidth]{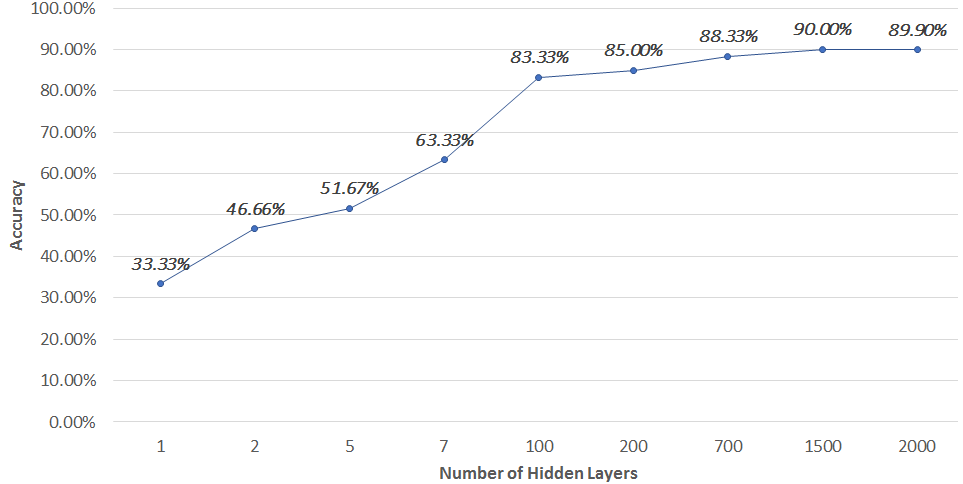}
\caption{Test Accuracy vs Number of Hidden Layers.}
\label{fig:exp1}
\end{figure}

For $HN \geq 100$, accuracy didn't improve much, but best accuracy achieved was 90\% when $HN = 1500$. In Fig.~\ref{fig:exp1}, it is clear that accuracy line is almost parallel to the x-axis from $HN = 100$ to $HN = 2000$. We can verify that accuracy of the neural network to correctly classify improves with an increase in the number of hidden layers \cite{surkan1990neural}.

\subsection{Experiment \#2}

\begin{table}[!htb]
\caption{Accuracy vs Number of Words selection}
\centering
\begin{tabular}{c|c|c}
\hline
{\textbf{Number of Words}} &{\textbf{No. of Hidden Layers}} & {\textbf{Accuracy}} \\ \hline
5500               & 100   & 81.67\%  \\
12000              & 100   & 88.33\%  \\
 \hline
\end{tabular}
\label{table:numberofwords}
\end{table}

In this model of ANNs (Artificial Neural Network), English helping verbs and conjunction words such as ’and’, ’what’, ’of’ etc. have been excluded from the data stream as such words won’t add much in improving the accuracy of the classification model. Moreover, such words may deviate the actual results. The second experiment has been performed to see, how the algorithm performs when a number of words feed into ANNs vary. It can be noted from Table \ref{table:numberofwords} that accuracy of the model is improved by 6.67\% when $num\_words$ (number of words) value is increased by 118\%.
It is understood from the test that when a large number of words is feed into the neural network, model prediction accuracy increased. \\

\begin{figure}[!htb]
\centering
\includegraphics[width=0.8\textwidth]{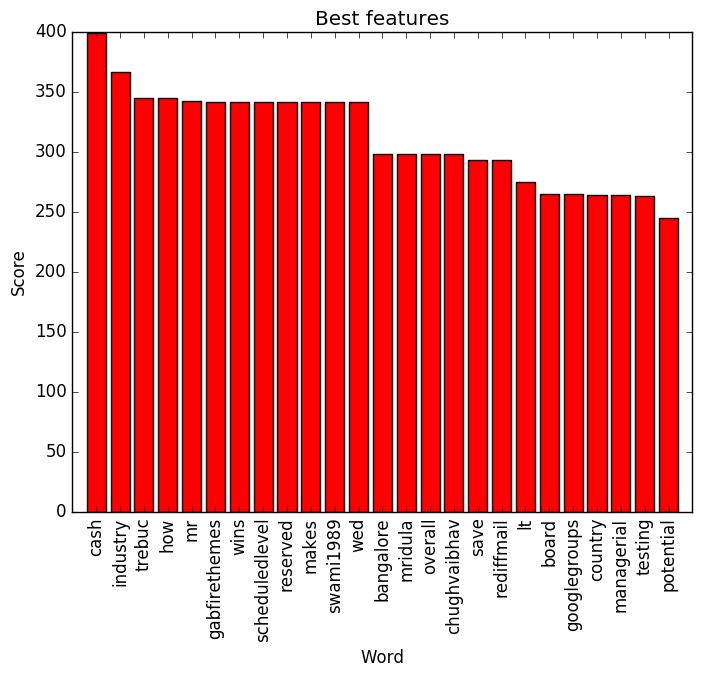}
\caption{Best Words vs Score.}
\label{fig:exp1}
\end{figure}

This model selects the best feature words based on chi-square using Scikit library. Word features selection allow the model to exclude less significant words from dataset to improve model correctness and data processing time, as the model is configured to not to include words which are not significant.

\section{Conclusions}

\begin{figure}[!htb]
\centering
\includegraphics[width=0.7\linewidth]{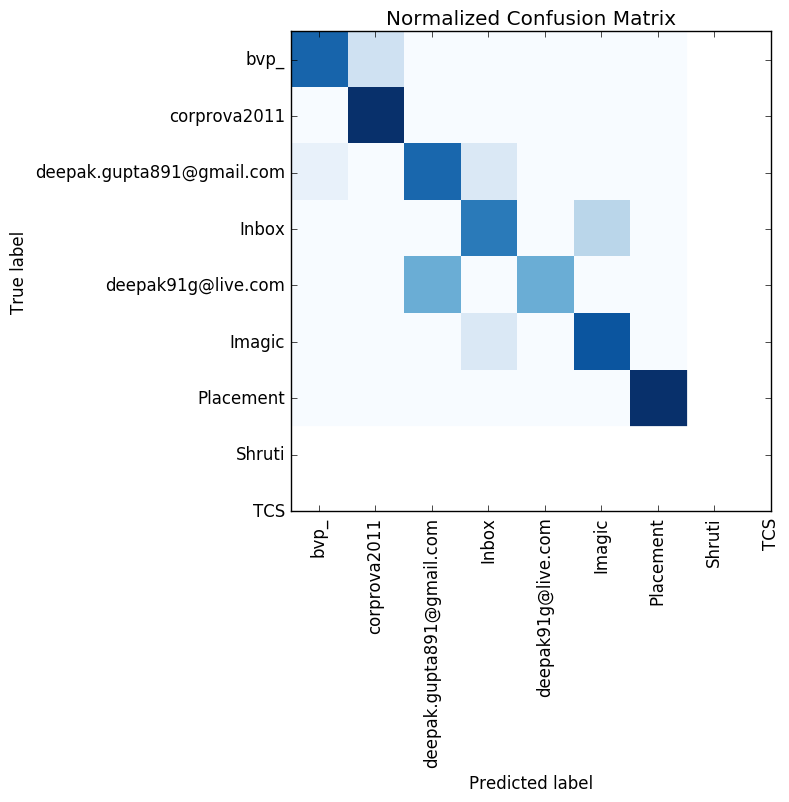}
\captionof{figure}{Confusion Matrix for Exp}
\label{fig:test1}
\end{figure}

During experiments, it is noted that more words in an email lead to better accuracy while keeping algorithm processing time lower. Datasets don't have enough number of email labelled for 'Friends' and 'Jobs'. From confusion matrix \cite{csurka2004visual} in Fig.~\ref{fig:test1}, it is seen that the model able to accurately classify labels for all email categories except for labels 'Friends' and 'Jobs'. After conducting two experiments, it is concluded that large dataset is required to train the model for classifying emails into the folder with high accuracy and model trained with more words selection has higher accuracy compared to the model which is trained with less number of words. \\

The diagonal line in Fig.~\ref{fig:test1} has dark coloured patches mean that the true positive value of each label is accurately classified by the model. \\

One can further improvise this algorithm by customising it for particular use case scenarios such us Customer Support, CEO Email Management, Enterprise level user etc.

%Figs.~\ref{code:wordcount} and \ref{fig:exp1}, and Table
%\ref{table:params}.

\section{Acknowledgements}
The first author would like to thank Prof Michael O'Neill, Dr Mike Fenton, Dr David Fagan for their help and support.
\bibliographystyle{abbrv}
\bibliography{biblio}

\end{document}